%% file: main.tex
\documentclass[letterpaper, 10 pt, conference]{ieeeconf}  %

\IEEEoverridecommandlockouts                              %

\input{macros/macros.tex}

\usepackage{graphicx}
\usepackage{amsmath}
\usepackage{bbm}
\usepackage{dsfont}
\usepackage{subcaption}

\PassOptionsToPackage{dvipsnames}{xcolor}
\PassOptionsToPackage{table}{xcolor}

\usepackage{amssymb}
\usepackage{booktabs}
\usepackage{makecell}
\usepackage[ruled,noend,linesnumbered]{algorithm2e}
\usepackage{titlesec}
\usepackage{wrapfig}
\usepackage{array}
\usepackage{multirow}
\usepackage{colortbl}
\usepackage{tabularx}

\usepackage{grffile}
\usepackage{xspace}
\usepackage{needspace}

\usepackage{comment} 
\usepackage{url}
\usepackage{bm}
\usepackage{paralist}

\usepackage{pifont}

\usepackage[belowskip=0pt,aboveskip=5pt,font=small]{caption}
\usepackage[belowskip=0pt,aboveskip=0pt,font=small]{subcaption}
\usepackage{xcolor}
\PassOptionsToPackage{hyphens}{url}
\usepackage[hidelinks]{hyperref}
\usepackage{xfrac}
\usepackage{cite}

\usepackage{caption}
\usepackage{wrapfig}

\def\eg{\emph{e.g.}}

\title{\LARGE \bf
FiLM-Nav: Efficient and Generalizable Navigation \\ via VLM Fine-tuning
}

\author{
    Naoki Yokoyama$^{1}$,
    Sehoon Ha$^{1}$
}

\begin{document}

\twocolumn[{
  \renewcommand\twocolumn[1][]{#1}
  \maketitle
  \centering
  \includegraphics[width=\textwidth,trim={0 2.14in 1.43in 0},clip]{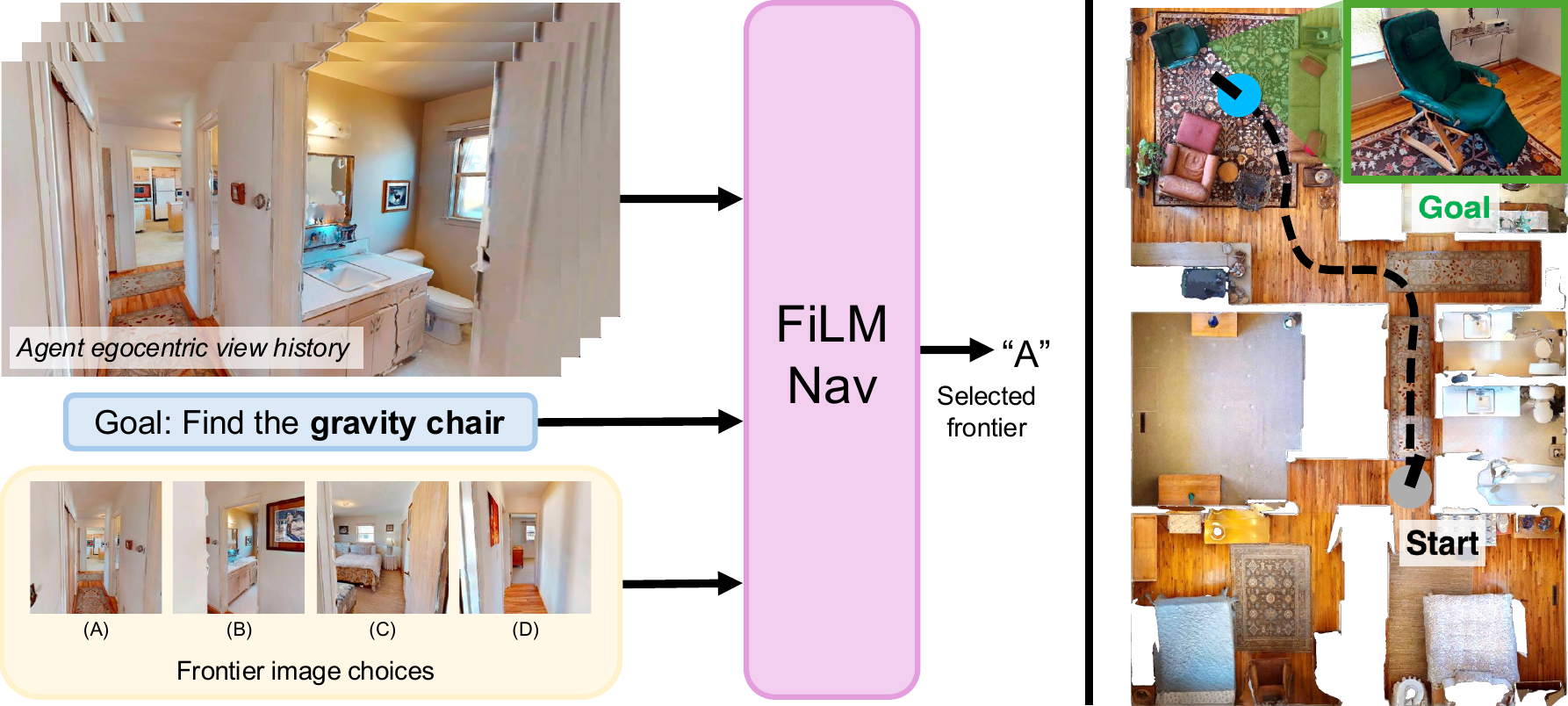}
  \captionof{figure} {
    \llmapproach takes the agent's egocentric view history, a language goal (\eg, ``Find the gravity chair"), and candidate frontier images shown on the left as input. The fine-tuned VLM policy then directly selects the next best frontier (``A") to explore. This process leads to efficient navigation paths and successful localization of target objects, as demonstrated on the right, showcasing strong performance and generalization even for object categories unseen during fine-tuning by drawing upon the vast world knowledge embedded within the pre-trained foundation model.
  }
  
}]

\footnotetext[1]{NY and SH are with Georgia Institute of Technology
{\tt\footnotesize nyokoyama@gatech.edu}}

\thispagestyle{empty}
\pagestyle{empty}

\begin{abstract}
Enabling robotic assistants to navigate complex environments and locate objects described in free-form language is a critical capability for real-world deployment. While foundation models, particularly Vision-Language Models (VLMs), offer powerful semantic understanding, effectively adapting their web-scale knowledge for embodied decision-making remains a key challenge. We present FiLM-Nav (\textbf{Fi}ne-tuned \textbf{L}anguage \textbf{M}odel for \textbf{Nav}igation), an approach that directly fine-tunes a pre-trained VLM as the navigation policy. In contrast to methods that use foundation models primarily in a zero-shot manner or for map annotation, FiLM-Nav learns to select the next best exploration frontier by conditioning directly on raw visual trajectory history and the navigation goal. Leveraging targeted simulated embodied experience allows the VLM to ground its powerful pre-trained representations in the specific dynamics and visual patterns relevant to goal-driven navigation. Critically, fine-tuning on a diverse data mixture combining ObjectNav, OVON, ImageNav, and an auxiliary spatial reasoning task proves essential for achieving robustness and broad generalization.
FiLM-Nav sets a new state-of-the-art in both SPL and success rate on HM3D ObjectNav among open-vocabulary methods, and sets a state-of-the-art SPL on the challenging HM3D-OVON benchmark, demonstrating strong generalization to unseen object categories.
Our work validates that directly fine-tuning VLMs on diverse simulated embodied data is a highly effective pathway towards generalizable and efficient semantic navigation capabilities.
\end{abstract}

\section{Introduction}
\label{sec:introduction}

The development of intelligent embodied agents capable of operating autonomously in complex, human-centric environments remains a central challenge in robotics and artificial intelligence. A critical capability for such agents is semantic navigation: the ability to understand high-level goals, often specified in natural language, and navigate purposefully to find relevant objects or locations within previously unseen spaces. Object Goal Navigation (ObjectNav) \cite{batra2020objectnav} formalizes this challenge, tasking an agent with finding an instance of a specified object category (\eg, ``find a chair"). The recent extension to Open-Vocabulary ObjectNav (OVON) \cite{yokoyama2024ovon} further raises the bar, requiring agents to generalize to potentially any object described using free-form language, significantly increasing the task's complexity and real-world applicability.

Large-scale foundation models, particularly Vision-Language Models (VLMs) and Large Language Models (LLMs), have catalyzed significant progress in this domain. These models possess remarkable zero-shot reasoning and world knowledge capabilities \cite{bommasani2021opportunities}, which researchers have leveraged for embodied navigation. Existing approaches often use foundation models for high-level planning, instruction interpretation, reasoning about object locations based on semantic cues, or enhancing environmental representations \cite{zhou2023esc, yu2023l3mvn, nie2025wmnav, yokoyama2024vlfm, yuan2024gamap, long2024instructnav, ziliotto2025tango, yu2024vlngame}. However, these models, typically trained on disembodied web data, exhibit limitations in spatial-temporal reasoning and physical grounding crucial for efficient robotic task execution \cite{Chen_2024_CVPR, liu2024aligning}. Many methods maintain a separation between the foundation model and the low-level navigation policy, often relying on intermediate abstractions that may introduce bottlenecks. Effectively adapting these powerful models to the specific constraints of embodied decision-making remains a key challenge.

We introduce \llmapproachfull (\llmapproach), an approach that directly fine-tunes a pre-trained VLM \cite{zhao2025cobra} to serve as the core navigation policy. Instead of employing foundation models primarily for high-level planning, we formulate navigation as a sequential decision-making problem where the VLM selects the next best exploration frontier. Critically, \llmapproach processes sequences of raw egocentric visual observations representing the agent's trajectory history, bypassing the need for explicit semantic maps containing object labels or pre-computed features. This allows the VLM to implicitly learn relevant spatio-temporal representations and ground its powerful pre-trained knowledge in the specific dynamics relevant to goal-driven navigation. By fine-tuning on a diverse mixture of ObjectNav, OVON, and ImageNav \cite{zhu2017target} tasks, and conditioning the policy on past visual experience, \llmapproach effectively adapts the VLM's pre-trained knowledge for embodied control, achieving richer grounding and avoiding potential information bottlenecks associated with intermediate representations.

Our primary contributions are:
(1) the development of \llmapproach, demonstrating the effectiveness of directly fine-tuning a VLM as a frontier-based navigation policy using raw visual history;
(2) state-of-the-art SPL and success performance on HM3D ObjectNav~\cite{ramakrishnan2021hm3d} among open-vocabulary methods, and state-of-the-art SPL on the challenging HM3D-OVON benchmark~\cite{yokoyama2024ovon}, demonstrating strong generalization to unseen object categories;
and (3) an ablation study validating the crucial role of diverse task training for robust and efficient navigation performance.

\section{Related Work}
\label{sec:related_work}

Foundation models (VLMs and LLMs) are becoming increasingly leveraged for ObjectNav \cite{batra2020objectnav} and OVON \cite{yokoyama2024ovon}. Prior research integrating these models can be broadly categorized based on their role in the navigation pipeline.
One major line of work utilizes foundation models primarily for high-level reasoning and planning. Methods like ESC \cite{zhou2023esc}, L3MVN \cite{yu2023l3mvn}, InstructNav \cite{long2024instructnav}, and WMNav \cite{nie2025wmnav} typically employ LLMs or VLMs to interpret complex instructions, infer likely object locations based on semantic knowledge (\eg, ``sofas are usually in living rooms"), or generate sequences of sub-goals or waypoints. These high-level plans are then often used to guide more traditional exploration or navigation policies (such as frontier-based exploration) operating on geometric or semantic maps. In these approaches, the foundation model generally remains separate from the low-level control policy and often relies on converting rich visual information into more abstracted representations (textual descriptions, symbolic states, map annotations) suitable for the language model's input space.

Another line of work focuses on enhancing the environmental representation itself using foundation models. VLFM, \cite{yokoyama2024vlfm}, TANGO \cite{ziliotto2025tango}, and VLN-Game \cite{yu2024vlngame} generate language-grounded value maps, with GAMap \cite{yuan2024gamap} also incorporating geometric parts and affordances. These methods seek richer representations than simple categorical maps that can be used to plan more efficient paths. Some mapless approaches also exist, such as ImagineNav \cite{zhao2025imaginenav}, which uses imagined future views to guide a VLM.

In contrast to prior work, \llmapproach directly fine-tunes a VLM \cite{zhao2025cobra} as the navigation policy to learn to select frontiers using raw visual history collected from diverse embodied data (ObjectNav, OVON, ImageNav \cite{zhu2017target}). Directly learning the navigation policy by fine-tuning the VLM on embodied experience, rather than using the foundation model solely for zero-shot high-level reasoning or as a separate map annotation module, allows for better grounding of world knowledge in the specific embodied task context compared to zero-shot inference or map annotation approaches. And unlike Uni-NaVid~\cite{zhang2024uninavid}, \llmapproach does so without learning an end-to-end policy that is conditioned on a particular embodiment or low-level action space.

\begin{figure*}[!t]
\centering
  \vskip5pt
  \includegraphics[width=\textwidth,trim={0 3.8in 2.01in 0},clip]{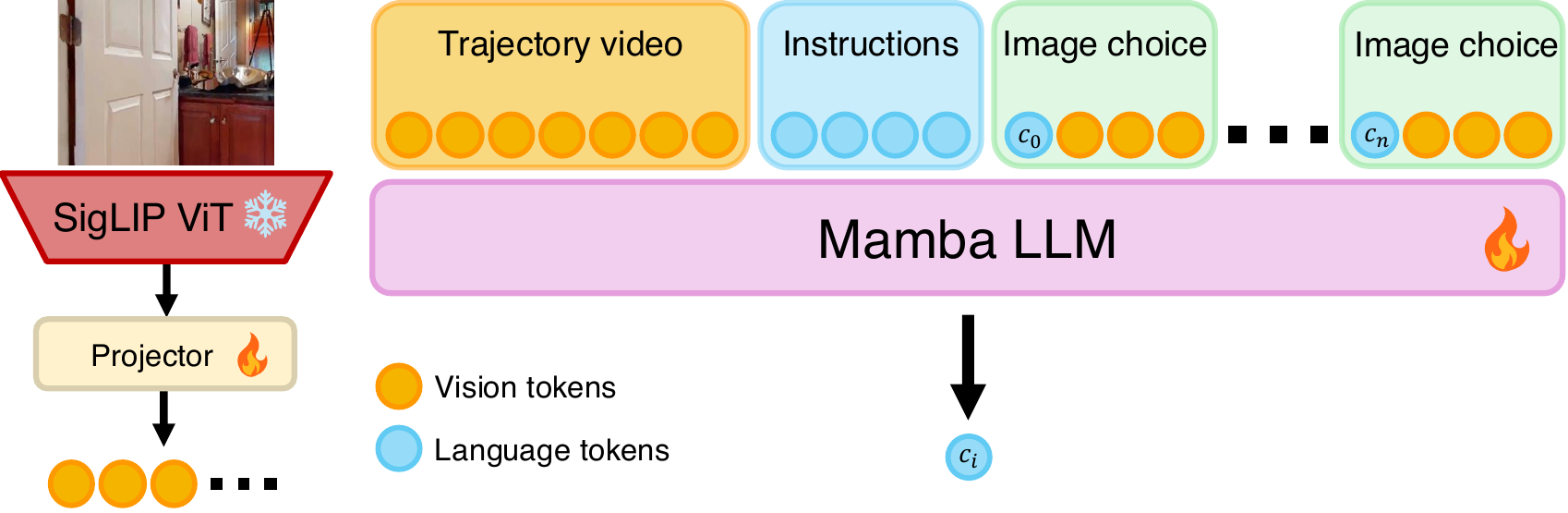}
  \captionof{figure} {
    \llmapproach architecture. \textit{Left}: Input images are processed into vision tokens using a frozen SigLIP ViT and a trainable projector. \textit{Right}: The LLM processes a sequence containing trajectory video tokens, language instructions, and image choices, each with vision tokens and a unique language token $c_i$. The LLM predicts the language token corresponding to the selected choice.
  }
  \label{fig:proposed_architecture}
\end{figure*}

\section{Problem Formulation}
\label{sec:problem_formulation}

Our paper addresses ObjectNav, where an embodied agent, placed in a previously unseen environment, must navigate to find an instance of a specific object category.

\textbf{Task Setup.} \label{sec:task_setup}
We follow the standard ObjectNav and OVON setup within HM3D environments \cite{ramakrishnan2021hm3d}. The agent uses egocentric RGB-D and an odometry sensor providing the agent's pose relative to its starting pose in the episode. At each step, it selects from a discrete set of actions: (\texttt{move\_forward(0.25m)}, \texttt{turn\_left(30\textdegree)},  \texttt{turn\_right(30\textdegree)}, \texttt{stop}). 
Given an object category specified via language (\eg, ``chair'', ``stove''), the agent must call \texttt{stop} within 1m of a target instance before exceeding 500 steps.

\textbf{Evaluation Metrics.} \label{sec:eval_metrics}
We evaluate navigation performance using two standard metrics \cite{anderson_arxiv18}: Success Rate (SR) and Success weighted by Path Length (SPL). SPL extends SR by incorporating path efficiency in addition to success, defined as:

\begin{equation}
\text{SPL} = \frac{1}{N} \sum_{i=1}^{N} S_i \frac{L_i}{\max{(P_i, L_i)}}
\end{equation}
where $S_i$ is the success indicator for episode $i$ ($0$ or $1$), $L_i$ is the shortest path length, and $P_i$ is the agent's actual path length. By penalizing unnecessarily long trajectories, SPL provides a more comprehensive assessment of navigation quality that balances task completion with efficiency, which is critical for practical robotic deployment where time, energy, and computational resources are constrained. Consequently, SPL serves as the primary ranking metric in Habitat Challenge competitions \cite{habitatchallenge2023}. Higher values indicate better performance for both metrics.

\section{Method: \llmapproachfull (\llmapproach)}
\label{sec:method}

We introduce \llmapproachfull (\llmapproach), a novel approach that directly fine-tunes a pre-trained Vision-Language Model (VLM) to act as a high-level navigation policy for ObjectNav and OVON tasks. Instead of relying on zero-shot prompting or intermediate map representations, \llmapproach learns to select promising exploration frontiers based directly on raw visual history and language goals.

\textbf{System Overview.} \label{sec:system_overview}
At each timestep during navigation, \llmapproach takes the agent's recent visual trajectory history, the natural language goal description, and visual representations of the current frontiers (boundaries between areas that have been explored and areas that have not been explored yet) as input. The core VLM processes this multi-modal sequence and outputs a choice corresponding to the frontier the agent should navigate towards. This high-level decision (frontier selection) passes a 2D waypoint to a low-level controller, which computes the next discrete action the robot should execute.

\textbf{VLM Architecture and Input Processing.} \label{sec:llm_architecture_method}
We utilize an efficient Mamba-based \cite{mamba} VLM architecture, specifically the 2.8 billion parameter Cobra model pre-trained on SlimPajamas \cite{cerebras2023slimpajama} and fine-tuned on web-scale image-text data \cite{zhao2025cobra}. Mamba's State Space Model (SSM) architecture offers computational advantages over traditional transformers, particularly linear scaling of memory and computation with sequence length, making it well-suited for processing long sequences of vision tokens inherent in visual navigation tasks. In our empirical measurements, Mamba provides 60\% faster inference (0.2s vs. 0.32s average) and 24\% lower memory usage (12.5GB vs. 15.5GB for inference) compared to similarly-sized transformer models. Unlike transformers with growing KV-caches, Mamba maintains constant memory usage throughout inference, enabling more frequent replanning and extended operational periods critical for robotic deployment. Additionally, Cobra demonstrated superior spatial reasoning capabilities on vision-language benchmarks \cite{zhao2025cobra}, outperforming larger 7B+ transformer models despite having only 2.8B parameters, which is particularly relevant for navigation tasks requiring understanding of spatial relationships.

For multi-modal processing, we employ a pre-trained SigLIP ViT \cite{zhai2023sigmoid} as the image encoder. Visual features extracted by the frozen SigLIP encoder are projected into the LLM's input embedding space using a trainable linear projection layer.

In order to handle long sequences of egocentric visual observations (trajectory history), we implement specific processing steps (Figure~\ref{fig:proposed_architecture}, left). Each frame is encoded into a grid of visual feature tokens. This grid undergoes average pooling across height, width, and temporal dimensions to create a compact representation before projection. For static images (\eg, frontier views, ImageNav goals), only height/width average pooling is applied before projection.

\begin{figure*}[!t]
\centering
  \vskip5pt
  \includegraphics[width=\textwidth,trim={0 2.87in 4.34in 0},clip]{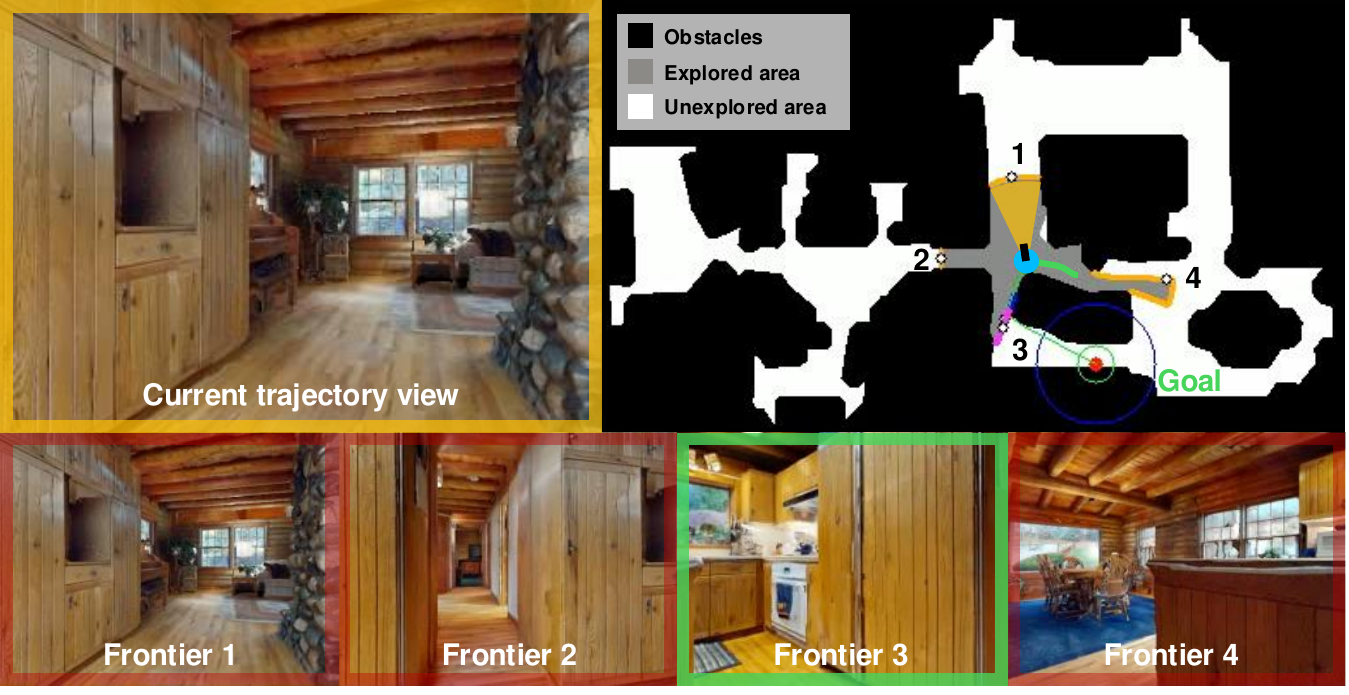}
  \captionof{figure} {
    Training data is generated via greedy frontier-based exploration. Each frontier is represented by a past RGB observation, selected from a pose offering optimal visibility towards the unexplored area. The frontier on the shortest path to the goal is recorded as the correct choice. \llmapproach uses these frontier images as goal-conditioned navigation options.
  }
  \label{fig:frontier_images}
\end{figure*}

\textbf{Input Sequence Formulation.} At each timestep $t$, the FiLM-Nav model receives a structured input sequence composed of the following token types: 
\textbf{(1) Trajectory History Video Tokens:} Spatio-temporally pooled tokens from recent egocentric RGB history (up to 20 frames). 
\textbf{(2) Task Instruction Tokens:} Language tokens specifying the navigation goal. 
\textbf{(3) ImageNav Goal Image Tokens (Conditional):} For ImageNav tasks only, height/width pooled vision tokens representing the goal image. 
\textbf{(4) Frontier Choice Tokens:} For each available frontier $k \in \{A, B, ..., N\}$, a pair comprising: (a) A unique language token for the choice label (\eg, `A'), and (b) height/width pooled vision tokens for the representative RGB image (detailed next) of frontier $k$.

A conceptual representation of the input sequence is: \\
\texttt{[<hist\_video\_tokens>] [<instr\_tokens>] [<opt\_imgnav\_goal>] [A] [<frontier\_A\_img>] [B] [<frontier\_B\_img>] ... [N] [<frontier\_N\_img>]}

The model's task is to predict the next token, which corresponds to the single language token (\eg, `A') identifying the chosen frontier.

\textbf{Frontier-Based Policy Learning.} \label{sec:frontier_policy}
We frame the navigation problem as learning a policy $\pi(f_t | H_t, g)$ that selects the next frontier $f_t$ to explore, conditioned on the trajectory history $H_t$ (represented by pooled vision tokens) and the goal $g$ (represented by language tokens, or potentially image tokens for ImageNav). The VLM directly embodies this policy, learning to output the token corresponding to the optimal frontier.

\textbf{Navigation Task Training Data Generation.} \label{sec:dataset_construction}
To fine-tune \llmapproach, we generate a large-scale dataset of supervised navigation examples using the Habitat simulator \cite{habitat19iccv}. We collect trajectories for ObjectNav, OVON, and Image Goal Navigation (ImageNav) \cite{zhu2017target} within their respective training scene splits. In ImageNav, the agent must navigate to a location specified by a goal image rather than a language description. Incorporating ImageNav significantly broadens the environmental diversity experienced during fine-tuning; since object annotations are not required to generate image goals, the entire training set of 3D scenes from the HM3D dataset can be utilized (800 training scenes for ImageNav versus 145 for ObjectNav/OVON).

For all tasks, instead of using simple shortest paths, trajectories are generated using a greedy frontier-based exploration oracle \cite{yamauchi1997frontier} combined with goal-seeking behavior. This approach prevents the model from overfitting to overly direct shortest paths, encouraging it to learn exploration behaviors needed to find targets that are not initially visible, a known limitation when training solely on optimal paths \cite{yokoyama2024ovon, ramrakhya2023pirlnav}. An occupancy map is built incrementally using depth and odometry, and frontiers (boundaries between explored and unexplored space) are identified. 
We design the greedy oracle to normally navigate towards the closest frontier, but to then switch to navigating towards the frontier leading to the goal once a goal instance (or the ImageNav goal location) is nearby (\beelineDistance meters).

At each timestep during trajectory generation, we record the agent's egocentric RGB-D observation, the set of all currently identified frontiers, and the `correct' frontier label (defined as the frontier lying on the shortest path to the closest goal instance).

Each frontier is visually represented by an RGB image selected from the agent's history (Figure~\ref{fig:frontier_images}). Importantly, frontier observations are collected from the robot's past trajectory history, \textit{not from the frontier locations themselves}. The image chosen corresponds to the camera pose that maximizes perpendicularity between the camera's view direction and the boundary constituting the frontier, providing the best view \textit{toward} the unexplored area. This ensures no training-inference gap, as the agent only uses observations it has already captured during the current episode, with the same frontier representation process operating identically during both training and evaluation.

\textbf{Auxiliary Spatial Reasoning Task.} \label{sec:aux_task}
To explicitly enhance the model's spatial reasoning, crucial for efficient navigation, we introduce an auxiliary training task. During trajectory generation within training environments, we collect the agent's recent visual history (last 20 frames) and capture several candidate RGB images from nearby locations visited during the segment. At the trajectory's end, the model receives: (1) its egocentric visual history as video tokens, (2) multiple candidate images with unique letter identifiers (A, B, C, ...), and (3) a multiple-choice prompt formatted as ``Which part of the environment is located at (X,Y)?" where (X,Y) denotes a 2D location relative to the robot's local frame at the current timestep. The model must predict the letter token corresponding to the correct candidate image, which is the image that taken from the given relative coordinates. This task forces the model to learn correspondences between first-person visual experience over time, and infer how different spaces are arranged in typical real-world indoor environments.

\textbf{Fine-tuning Procedure.} \label{sec:finetuning}
We fine-tune the pre-trained VLM using the combined dataset from ObjectNav, OVON, ImageNav, and the auxiliary spatial task. The training objective is the standard next-token prediction loss (cross-entropy) commonly used for LLMs. The model is trained to predict the single token corresponding to the correct frontier choice (or the correct image choice for the auxiliary task). During fine-tuning, the weights of the SigLIP ViT image encoder are kept frozen while we train the linear projection layer and the LLM weights.

We use the AdamW optimizer with learning rate $2 \times 10^{-5}$, global batch size of 576 (achieved through distributed training and gradient accumulation), and train for 3 epochs on 8 A40 GPUs. The combined dataset comprises approximately 26k samples each for ObjectNav and OVON, 40k for ImageNav, and 30k for the auxiliary spatial task. Training takes around 15 hours total.

\textbf{Online Deployment within Habitat.} \label{sec:online_deployment}
During evaluation, \llmapproach operates in a closed loop with the following components: (1) \textit{Mapping and Frontier Extraction}: An occupancy map is updated using depth and odometry; frontiers are extracted as boundaries between explored/unexplored areas. (2) \textit{Frontier Representation}: Representative RGB images for each frontier are selected from the observation history by choosing past camera poses that maximize perpendicularity with frontier boundaries. (3) \textit{VLM Policy Inference}: The trajectory history, language goal, and frontier images are formatted into the input sequence; the fine-tuned VLM predicts the optimal frontier token. (4) \textit{Low-Level Control}: The coordinates of the chosen frontier are passed to a pre-trained PointNav policy \cite{wijmans2019dd} to predict the next discrete action to take. (5) \textit{Object Detection}: similar to \cite{yokoyama2024vlfm}, we use OWLViTv2 \cite{minderer2023scaling} to detect the goal object, and upon detection, MobileSAM \cite{mobile_sam} is used to segment the object's point cloud from the depth image. The PointNav policy takes as input the point on the cloud that is closest to the agent, and \texttt{stop} is called once the agent is within the success distance.

\begin{table}[!t]
    \centering
    \vskip5pt
    \caption{Performance on the HM3D-v0.1 ObjectNav validation set. Methods that can only support the six object categories present in the HM3D ObjectNav benchmark are grayed out. \llmapproach achieves the highest SPL and the highest SR among methods that are open-vocabulary.}
    \label{tab:llm_hm3d_results}
    \resizebox{\columnwidth}{!}{
        \begin{tabular}{@{}lcrr@{}}
            \toprule
            Method & Open-vocab? & SR $(\uparrow)$ & SPL $(\uparrow)$ \\
            \midrule
            \textcolor{gray}{OVRLv2 \cite{yadav2023ovrl}} & \textcolor{gray}{\ding{55}} & \textcolor{gray}{$64.7\%$} & \textcolor{gray}{$28.1\%$} \\
            \textcolor{gray}{PIRLNav \cite{ramrakhya2023pirlnav}} & \textcolor{gray}{\ding{55}} & \textcolor{gray}{$\mathbf{70.4}\%$} & \textcolor{gray}{$34.1\%$} \\
            \midrule
            GAMap \cite{yuan2024gamap} & \checkmark & $53.1\%$ & $26.0\%$ \\
            VLN-Game \cite{yu2024vlngame} & \checkmark & $61.3\%$ & $26.8\%$ \\
            VLFM \cite{yokoyama2024vlfm} & \checkmark & $52.5\%$ & $30.4\%$ \\
            WMNav \cite{nie2025wmnav} & \checkmark & $58.1\%$ & $31.2\%$ \\
            \midrule
            \llmapproach (Ours) & \checkmark & $\mathbf{61.7}\%$ & $\mathbf{37.3}\%$ \\
            \bottomrule
        \end{tabular}
    }
\end{table}

\section{Experiments}
\label{sec:experiments}

In this section, we detail the experimental setup designed to evaluate the performance of \llmapproach and validate our contributions. We aim to answer the following key research questions:
\vspace{-0.5em}
\begin{enumerate}
    \item How does \llmapproach perform on standard HM3D ObjectNav benchmarks compared to other state-of-the-art methods capable of open-vocabulary navigation?
    \item How effectively does \llmapproach generalize to the challenges of Open-Vocabulary ObjectNav (OVON), particularly regarding unseen object categories and synonyms?
    \item What are the contributions of the different components of our training strategy (diverse task mixture, auxiliary spatial task) to the overall performance?
\end{enumerate}

\textbf{Experimental Setup.}
All experiments are conducted using the Habitat simulator \cite{habitat19iccv} with 3D scans from the HM3D dataset \cite{ramakrishnan2021hm3d}. We follow the standard ObjectNav task setup as described in Section~\ref{sec:problem_formulation}.

\textbf{Tasks and Benchmarks.}
We evaluate \llmapproach on standard benchmarks: 
\textbf{HM3D ObjectNav}, which uses a predefined set of six object categories, using the validation sets of HM3D-v0.1 \cite{ramakrishnan2021hm3d} and the improved HM3D-v0.2; 
and \textbf{HM3D-OVON} \cite{yokoyama2024ovon}, which tests generalization via its validation splits: \vseen (categories seen during OVON training), \vue (unseen synonyms/neighbors of seen categories), and \vuh (semantically distinct unseen categories). Performance on the OVON unseen splits, particularly \vue and \vuh, directly measures generalization to novel language descriptions of objects.

\textbf{Baselines.}
We compare \llmapproach against recent state-of-the-art methods.
For \textbf{HM3D ObjectNav}, baselines include GAMap~\cite{yuan2024gamap}, VLN-Game~\cite{yu2024vlngame}, VLFM~\cite{yokoyama2024vlfm}, WMNav~\cite{nie2025wmnav}, ImagineNav~\cite{zhao2025imaginenav}, InstructNav~\cite{long2024instructnav}, and Uni-NaVid~\cite{zhang2024uninavid} (selecting strongest open-vocabulary methods available on respective benchmark versions v0.1/v0.2). 
For \textbf{HM3D-OVON}, baselines are VLFM~\cite{yokoyama2024vlfm}, DAgRL+OD~\cite{yokoyama2024ovon}, and TANGO~\cite{ziliotto2025tango}.

\section{Results and Analysis}
\label{sec:results}

In this section, we present the empirical evaluation of \llmapproach (\llmapproachfull).

\textbf{Performance on HM3D ObjectNav.}
We first evaluate \llmapproach on the standard HM3D ObjectNav validation sets, comparing against state-of-the-art methods, with a focus on those that are also capable of open-vocabulary navigation (versus those that only support the six categories that appear in the HM3D ObjectNav dataset). Table~\ref{tab:llm_hm3d_results} shows the results on the HM3D-v0.1 benchmark, and Table~\ref{tab:llm_hm3d_v2_results} presents the results on the improved HM3D-v0.2 benchmark.

\begin{table}[!t]
    \centering
    \vskip5pt
    \caption{Performance on the HM3D-v0.2 ObjectNav validation set. \llmapproach sets a new state-of-the-art in both SR and SPL.}
    \label{tab:llm_hm3d_v2_results}
    \resizebox{\columnwidth}{!}{
        \begin{tabular}{@{}lcrr@{}}
            \toprule
            Method & Open-vocab? & SR $(\uparrow)$ & SPL $(\uparrow)$ \\
            \midrule
            ImagineNav \cite{zhao2025imaginenav} & \checkmark & $53.0\%$ & $23.8\%$ \\
            InstructNav \cite{long2024instructnav} & \checkmark & $58.0\%$ & $20.9\%$ \\
            VLFM \cite{yokoyama2024vlfm} & \checkmark & $62.6\%$ & $31.0\%$ \\
            WMNav \cite{nie2025wmnav} & \checkmark & $72.2\%$ & $33.3\%$ \\
            Uni-NaVid \cite{zhang2024uninavid} & \checkmark & $73.7\%$ & $37.1\%$ \\
            \midrule
            \llmapproach (Ours) & \checkmark & $\mathbf{77.0}\%$ & $\mathbf{41.3}\%$ \\
            \bottomrule
        \end{tabular}
    }
\end{table}

On the HM3D-v0.1 benchmark (Table~\ref{tab:llm_hm3d_results}), \llmapproach achieves the highest Success weighted by Path Length (SPL) score of $37.3\%$. This significantly surpasses the previous best SPL among compared open-vocabulary methods (WMNav \cite{nie2025wmnav} with $31.2\%$), representing an absolute improvement of $6.1\%$ and a relative improvement of $19.6\%$. Notably, it also surpasses PIRLNav \cite{ramrakhya2023pirlnav} (34.1\% SPL) by 3.2\% absolute and 9.4\% relative, despite PIRLNav being trained using nearly 2,400 hours of human demonstrations collected within Habitat specifically for the six standard ObjectNav categories. These results highlight the superior path efficiency achieved by directly fine-tuning the VLM as a policy. Our method also achieves a state-of-the-art success rate (SR) of $61.7\%$ among open-vocabulary methods.

It is worth noting that our current frontier-based mapping implementation operates in 2D, and cannot handle stair traversal due to the lack of z-coordinate information in the sensor configuration. On the other hand, methods like PIRLNav and OVRL learn end-to-end policies capable of multi-floor navigation, as they are directly trained to output discrete actions to search for the six goal object categories used by the benchmark without constructing maps. This accounts for failures in 13.9\% of HM3D-v0.1 episodes that require floor transitions to go from the start position to an instance of the goal category. Despite this limitation and our reliance on a pre-trained object detector (versus end-to-end object recognition training that adapts to HM3D-specific visual characteristics), \llmapproach's efficient path planning enables it to achieve superior SPL scores.

Although \llmapproach has a lower SR than PIRLNav ($61.7\%$ vs. $70.4\%$), it achieves a higher SPL score ($37.3\%$ vs. $34.1\%$) due to its exceptional path efficiency. Crucially, SPL measures ``efficient success''; since unsuccessful episodes contribute 0 to SPL, achieving \textit{higher} SPL despite \textit{lower} SR means \textit{\llmapproach's successful episodes are dramatically more efficient}. This demonstrates that when \llmapproach successfully reaches the goal, it does so via substantially shorter paths than competing methods.

On the HM3D-v0.2 benchmark (Table~\ref{tab:llm_hm3d_v2_results}), \llmapproach establishes new state-of-the-art performance in both metrics, achieving $77.0\%$ SR and $41.3\%$ SPL. It outperforms the prior leading method, Uni-NaVid \cite{zhang2024uninavid} ($73.7\%$ SR / $37.1\%$ SPL), by $3.3\%$ absolute SR and $4.2\%$ absolute SPL (an $11.3\%$ relative SPL improvement). The strong performance on both benchmarks, particularly the substantial gains in SPL, demonstrates the effectiveness of our fine-tuning approach for generating efficient navigation behaviors in standard ObjectNav tasks.

\textbf{Performance on Open-Vocabulary ObjectNav (OVON).}
Next, we evaluate \llmapproach on the more challenging HM3D-OVON benchmark \cite{yokoyama2024ovon}, designed to test generalization to a wide range of object categories beyond those seen during standard ObjectNav training. Table~\ref{tab:llm_ovon_results} presents the results across the three standard OVON validation splits: \vseen (seen categories), \vue (unseen synonyms/close categories), and \vuh (unseen distinct categories).

\newcommand{\ovoncolsep}{\hspace{4pt}}
\begin{table}[!t]
    \centering
    \vskip5pt
    \caption{Performance comparison on the HM3D-OVON validation splits. \llmapproach achieves state-of-the-art performance in most metrics across the splits, with the highest SPL values throughout.}
    \label{tab:llm_ovon_results}
    \resizebox{\columnwidth}{!}{
        \begin{tabular}{@{}lc@{\ovoncolsep}cc@{\ovoncolsep}cc@{\ovoncolsep}c@{\ovoncolsep}}
            \toprule
            \multirow{2}{*}{Method} & \multicolumn{2}{c}{\textsc{Val Seen}} & \multicolumn{2}{c}{\textsc{Synonyms}} & \multicolumn{2}{c}{\textsc{Val Unseen}} \\
            \cmidrule(lr){2-3}\cmidrule(lr){4-5}\cmidrule(lr){6-7}
             & SR$\uparrow$ & SPL$\uparrow$ & SR$\uparrow$ & SPL$\uparrow$ & SR$\uparrow$ & SPL$\uparrow$ \\
            \midrule
            VLFM \cite{yokoyama2024vlfm} & $38.1$ & $20.8$ & $37.7$ & $21.2$ & $38.5$ & $22.2$ \\
            DAgRL+OD \cite{yokoyama2024ovon} & $38.5$ & $21.1$ & $39.0$ & $21.4$ & $37.1$ & $19.9$ \\
            TANGO \cite{ziliotto2025tango} & - & - & - & - & $35.5$ & $19.5$ \\
            Uni-NaVid \cite{zhang2024uninavid} & $41.3$ & $21.1$ & $\mathbf{43.9}$ & $21.8$ & $39.5$ & $19.8$ \\
            \midrule
            \llmapproach (Ours) & $\mathbf{44.9}$ & $\mathbf{24.5}$ & $40.1$ & $\mathbf{23.1}$ & $\mathbf{40.8}$ & $\mathbf{24.4}$ \\
            \bottomrule
        \end{tabular}
    }
\end{table}

Notably, \llmapproach achieves state-of-the-art SPL on all three splits of the OVON dataset, and state-of-the-art SR on two of the three.
On the \vseen split, \llmapproach obtains $44.9\%$ SR and $24.5\%$ SPL, outperforming the prior best (Uni-NaVid \cite{zhang2024uninavid}) by a significant margin ($+3.6\%$ absolute SR, $+3.4\%$ absolute SPL).
On the \vue split, which tests robustness to linguistic variations for known object types, Uni-NaVid achieves the highest SR at $43.9\%$, while \llmapproach achieves the second highest SR at $40.1\%$ and maintains the best SPL at $23.1\%$ compared to Uni-NaVid's $21.8\%$ SPL.
Crucially, on the \vuh split, which directly measures generalization to object categories unseen and unrelated to those seen during navigation training, \llmapproach achieves $40.8\%$ SR and $24.4\%$ SPL. This substantially surpasses the previous best results on this challenging split (Uni-NaVid \cite{zhang2024uninavid} with $39.5\%$ SR / $19.8\%$ SPL, and VLFM \cite{yokoyama2024vlfm} with $38.5\%$ SR / $22.2\%$ SPL). This strong performance on unseen categories underscores the benefit of fine-tuning a large pre-trained VLM; the model effectively leverages its broad world knowledge acquired during pre-training and adapts it through our navigation-focused fine-tuning process to successfully locate objects it has never been explicitly trained to navigate towards, suggesting that direct grounding via fine-tuning allows for more effective transfer of pre-trained knowledge to novel embodied scenarios compared to zero-shot methods or those relying on intermediate representations.
Overall, the results on OVON demonstrate that \llmapproach excels in standard ObjectNav and possesses strong generalization capabilities for open-vocabulary scenarios, achieving efficient and successful navigation for a diverse and extensive range of object goals described in language, particularly showing strength in SPL metrics across all splits.

\newcommand{\firstfourspacing}{0.3em}    %
\newcommand{\srpplspacing}{0.3em}        %
\newcommand{\groupspacing}{1.2em}        %

\begin{table}[!t]
    \centering
    \vskip5pt
    \caption{Ablation study evaluating contribution of training data sources (ON=ObjectNav, IN=ImageNav, OV=OVON) and the auxiliary task (AX) on HM3D-v0.2 and HM3D-OVON validation sets. Full model (bottom row) performs best.}
    \label{tab:llm_ablation_results}
    \resizebox{\columnwidth}{!}{
        \begin{tabular}{@{}c@{\hspace{\firstfourspacing}}c@{\hspace{\firstfourspacing}}c@{\hspace{\firstfourspacing}}c@{\hspace{1.0em}}c@{\hspace{\srpplspacing}}c@{\hspace{\groupspacing}}c@{\hspace{\srpplspacing}}c@{\hspace{\groupspacing}}c@{\hspace{\srpplspacing}}c@{\hspace{\groupspacing}}c@{\hspace{\srpplspacing}}c@{}}
            \toprule
            \multicolumn{4}{c}{} & \multicolumn{2}{c}{\textsc{HM3D}} & \multicolumn{2}{c}{\textsc{Val}} & \multicolumn{2}{c}{\textsc{Syno-}} & \multicolumn{2}{c}{\textsc{Val}} \\
            \multicolumn{4}{c}{Training data} & \multicolumn{2}{c}{\textsc{v0.2}} & \multicolumn{2}{c}{\textsc{Seen}} & \multicolumn{2}{c}{\textsc{nyms}} & \multicolumn{2}{c}{\textsc{Unseen}} \\
            \cmidrule(lr){1-4}\cmidrule(lr){5-6}\cmidrule(lr){7-8}\cmidrule(lr){9-10}\cmidrule(lr){11-12}
             ON & IN & OV & AX & SR & SPL & SR & SPL & SR & SPL & SR & SPL \\
            \midrule
            \checkmark & - & - & - & $74.3$ & $38.8$ & $41.7$ & $22.0$ & $38.8$ & $21.6$ & $39.4$ & $22.5$ \\
            \checkmark & \checkmark & - & - & $74.9$ & $39.1$ & $43.1$ & $22.8$ & $39.3$ & $21.7$ & $\mathbf{41.5}$ & $23.8$ \\
            \checkmark & \checkmark & \checkmark & - & $74.9$ & $39.2$ & $43.3$ & $23.3$ & $39.9$ & $22.4$ & $40.8$ & $23.6$ \\
            \midrule
            \checkmark & \checkmark & \checkmark & \checkmark & $\mathbf{77.0}$ & $\mathbf{41.3}$ & $\mathbf{44.9}$ & $\mathbf{24.5}$ & $\mathbf{40.1}$ & $\mathbf{23.1}$ & $40.8$ & $\mathbf{24.4}$ \\
            \bottomrule
        \end{tabular}
    }
\end{table}

\textbf{Ablation Study.} \label{sec:res_ablation}
To understand the contributions of the different components of our training strategy, we perform an ablation study investigating the impact of the training data mixture (ObjectNav, ImageNav, OVON datasets) and our proposed auxiliary spatial reasoning task. The results are presented in Table~\ref{tab:llm_ablation_results}, evaluated on HM3D-v0.2 and HM3D-OVON.

We start with a baseline model trained solely on the standard  ObjectNav dataset (Row 1). This model achieves strong performance on HM3D ($74.3\%$ SR / $38.8\%$ SPL) and exhibits considerable zero-shot generalization to OVON splits (\eg, $39.4\%$ SR / $22.5\%$ SPL on \vuh), again highlighting the benefit of the VLM's pre-trained knowledge in spite of the limited amount of categories encountered during navigation training.

Adding ImageNav training data (Row 2 vs. Row 1), which exposes the model to more diverse navigation scenarios across 800 scenes without object labels (vs. only 145 train scenes with labels), provides slight improvements, particularly boosting OVON generalization (\eg, \vuh SPL increases from $22.5\%$ to $23.8\%$). This improvement likely stems from ImageNav exposing the model to a vastly more diverse set of goal representations during training, since each unique goal image is projected into a unique language embedding using the SigLIP encoder and the projection layer, compared to the small, fixed set of embeddings corresponding to the 6 standard ObjectNav categories, thus enhancing the model's ability to generalize its goal-conditioning mechanism to the diverse set of possible goal object categories of OVON.

Incorporating OVON training data (Row 3 vs. Row 2), which includes explicit navigation examples for a wider variety of object categories, further improves performance on OVON splits (\vseen SPL from $22.8\%$ to $23.3\%$, \vue SPL from $21.7\%$ to $22.4\%$) while maintaining HM3D performance. This demonstrates the value of training on target distributions closer to the open-vocabulary evaluation.

Finally, incorporating our auxiliary spatial reasoning task (Row 4 vs. Row 3) provides further consistent gains, completing the full \llmapproach model. Performance improves across almost all metrics with a jump of up to $2.1\%$ in absolute SPL and $2.1\%$ in absolute SR (see Table~\ref{tab:llm_ablation_results}). This suggests that enhancing the model's ability to correlate visual history with spatial layout aids efficient frontier selection and contributes positively to overall navigation performance.

In summary, the ablation study confirms that leveraging a diverse mixture of navigation tasks (ObjectNav, ImageNav, OVON) and incorporating the targeted auxiliary spatial reasoning task are crucial components that contribute to the state-of-the-art performance of \llmapproach.

\textbf{Real-World Transferability.}
While our primary experiments utilize simulation for systematic evaluation, we provide evidence supporting transferability to real-world settings. First, multiple prior works have demonstrated successful \textit{zero-shot} sim-to-real transfer using models trained on the same HM3D dataset in Habitat~\cite{silwal2024learn, zhang2024uninavid}, validating the photorealistic simulation's effectiveness for real-world deployment. Second, \llmapproach operates at the frontier selection level of abstraction, avoiding low-level control details that typically cause sim-to-real failures in robotics. The model processes RGB images to select exploration directions, a capability that transfers more readily than precise motor control. Third, our use of a pre-trained VLM foundation provides inherent robustness to visual domain shift, as these models are trained on a web-scale dataset of diverse real-world imagery.

To empirically validate transferability, we conducted preliminary real-world evaluation: (1) We collected egocentric videos in a real apartment, with the camera trajectory capturing potential frontier directions. \llmapproach successfully selected correct frontiers toward goal objects when presented with these real-world observations. (2) We evaluated on randomly sampled POV tours from the RealEstate10K dataset~\cite{realestate10k}, labeling frames as frontiers and annotating whether they lead toward goal objects. The model demonstrated effective frontier selection on this real-world residential imagery. While deployment and configuration on a real mobile robot hardware platform remains to be future work, these results indicate that our approach can process and make decisions on real-world visual input effectively. Additional videos and evaluation details are available at \href{https://filmnav.netlify.app/}{\tt\textcolor{blue}{\underline{https://filmnav.netlify.app}}}.

\section{Conclusion}
\label{sec:conclusion}
This paper introduced \llmapproach, a novel approach that demonstrates the efficacy of directly fine-tuning a pre-trained VLM \cite{zhao2025cobra} as the core navigation policy for Object Goal Navigation (ObjectNav) and its open-vocabulary extension (OVON). Departing from methods that use foundation models primarily for zero-shot reasoning or map enhancement, \llmapproach formulates navigation as selecting the optimal exploration frontier conditioned directly on raw visual trajectory history and the language goal. Fine-tuning on a diverse mixture of simulated embodied tasks (ObjectNav, OVON, ImageNav) combined with an auxiliary spatial reasoning task effectively grounds the VLM's powerful pre-trained representations in the spatio-temporal context required for efficient, goal-directed navigation. Our experiments validate this strategy, showing that \llmapproach achieves state-of-the-art navigation efficiency (SPL) on standard HM3D ObjectNav benchmarks and establishes a new state-of-the-art in SPL across all validation splits of the demanding HM3D-OVON benchmark, which highlights its strong generalization to novel object categories.

While \llmapproach demonstrates strong performance, several limitations present opportunities for future work. The current framework targets finding \textit{any} instance of a specified category rather than specific instances with distinguishing properties, and focuses on single-object search tasks rather than complex multi-object instructions. Additionally, the system's reliance on an external object detector for termination and explicit occupancy mapping for frontier generation introduces dependencies that could be addressed through end-to-end learning. Despite these limitations, our work validates that directly fine-tuning VLMs on diverse simulated embodied data is a highly effective pathway toward generalizable and efficient semantic navigation capabilities, providing a foundation for developing even more versatile embodied agents.

\section{Acknowledgements}
\scriptsize{
    This work was supported by the Korea Evaluation Institute of Industrial Technology (KEIT) funded by the Korea Government (MOTIE) under Grant No.20018216, Development of Mobile Intelligence SW for autonomous navigation of legged robots in dynamic and atypical environments for real application.
}

\addtolength{\textheight}{-2cm}   %

\bibliographystyle{IEEEtran}
\bibliography{references}

\end{document}

%% file: macros/macros.tex
\makeatletter
\DeclareRobustCommand\onedot{\futurelet\@let@token\@onedot}
\def\@onedot{\ifx\@let@token.\else.\null\fi\xspace}

\def\eg{\emph{e.g}\onedot}

\makeatother

\newcounter{myenum}
	{\end{list}}

\usepackage{booktabs}       %
\usepackage{amsfonts}       %
\usepackage{nicefrac}       %
\usepackage{microtype}      %
\usepackage{colortbl}
\usepackage{placeins}

\usepackage{float}
\usepackage{fontawesome}
\usepackage{adjustbox}
\newcolumntype{R}[2]{%
    >{\adjustbox{angle=#1,lap=\width-(#2)}\bgroup}%
    l%
    <{\egroup}%
}

\usepackage{rotating}
\usepackage{wrapfig}
\usepackage{array}
\usepackage{multirow}
\usepackage{colortbl}
\usepackage{tabularx}
\usepackage{xfrac}

\usepackage{breqn}  %

\newcommand{\beelineDistance}{3.5\xspace}

\newcommand{\vseen}{\textsc{Val Seen}\xspace}
\newcommand{\vue}{\textsc{Val Seen Synonyms}\xspace}
\newcommand{\vuh}{\textsc{Val Unseen}\xspace}

\newcommand{\llmapproachfull}{\textbf{Fi}ne-tuned \textbf{L}anguage \textbf{M}odel for \textbf{N}avigation\xspace}
\newcommand{\llmapproach}{FiLM-Nav\xspace}